\documentclass[10pt,twocolumn,letterpaper]{article}

\usepackage{iccv}
\usepackage{times}
\usepackage{epsfig}
\usepackage{graphicx}
\usepackage{tikz}
\usepackage{amsmath}
\usepackage{amssymb}

\usetikzlibrary{shapes.geometric, arrows}
\tikzstyle{basic} = [rectangle, minimum width=2cm, minimum height=0.5cm, text centered, draw=black]

\tikzstyle{arrow} = [thick,->,>=stealth]

\usepackage{listings}
\usepackage{color}
\usepackage{psfrag}
\usepackage{bm}
\usepackage{subcaption}
\usepackage{booktabs}

\usepackage{romannum}
\usepackage[page]{appendix}

\usepackage[breaklinks=true,bookmarks=false]{hyperref}

\iccvfinalcopy % *** Uncomment this line for the final submission

\def\iccvPaperID{10937} % *** Enter the ICCV Paper ID here

\ificcvfinal\fi

\definecolor{mygreen}{rgb}{0,0.6,0}
\definecolor{mygray}{rgb}{0.5,0.5,0.5}
\definecolor{mymauve}{rgb}{0.58,0,0.82}

\begin{document}

%%%%%%%%% TITLE
\title{Tiled Squeeze-and-Excite: Channel Attention With Local Spatial Context}

% \author{Niv Vosco\\
% Hailo.ai\\
% {\tt\small nivv@hailo.ai}
% \and
% Alon Shenkler\\
% Hailo.ai\\
% {\tt\small }
% \and
% Mark Grobman\\
% Hailo\\
% {\tt\small markg@hailo.ai}
% }

\title{Tiled Squeeze-and-Excite: Channel Attention With Local Spatial Context}
\author{Niv Vosco \qquad Alon Shenkler \qquad Mark Grobman\\
Hailo \\\{nivv, markg\}@hailo.ai}
\date{\vspace{-5ex}}

\maketitle
\ificcvfinal\thispagestyle{empty}\fi

%%%%%%%%% ABSTRACT
\begin{abstract}
In this paper we investigate the amount of spatial context required for channel attention. To this end we study the popular squeeze-and-excite (SE) block which is a simple and lightweight channel attention mechanism. SE blocks and its numerous variants commonly use global average pooling (GAP) to create a single descriptor for each channel. Here, we empirically analyze the amount of spatial context needed for effective channel attention and find that limited local-context on the order of seven rows or columns of the original image is sufficient to match the performance of global context. We propose tiled squeeze-and-excite (TSE), which is a framework for building SE-like blocks that employ several descriptors per channel, with each descriptor based on local context only. We further show that TSE is a drop-in replacement for the SE block and can be used in existing SE networks without re-training. This implies that local context descriptors are similar both to each other and to the global context descriptor. Finally, we show that TSE has important practical implications for deployment of SE-networks to dataflow AI accelerators due to their reduced pipeline buffering requirements. For example, using TSE reduces the amount of activation pipeline buffering in EfficientDet-D2 by 90\% compared to SE (from 50M to 4.77M) without loss of accuracy. Our code and pre-trained models will be publicly available.  
\end{abstract}

%%%%%%%%% BODY TEXT

%-------------------------------------------------------------------------
%----------------------- Introduction ------------------------------------
%-------------------------------------------------------------------------

\section{Introduction} \label{intro_sec}

\begin{figure}
\centering
\psfrag{A}[cc][cc][0.5]{$C$}
\psfrag{B}[cc][cc][0.5]{$H$}
\psfrag{C}[cc][cc][0.5]{$W$}
\psfrag{D}[cc][cc][0.5]{$GAP$}
\psfrag{F}[cc][cc][0.5]{$Conv$}
\psfrag{G}[cc][cc][0.5]{${C/r}$}
\psfrag{J}[cc][cc][0.5]{$\times$}
\psfrag{T}[cc][cc][0.5]{$Pool$}
\psfrag{R}[cc][cc][0.5]{$Conv$}
\psfrag{X}[cc][cc][0.5]{$Buffer$}
\centering
\includegraphics[scale=0.24]{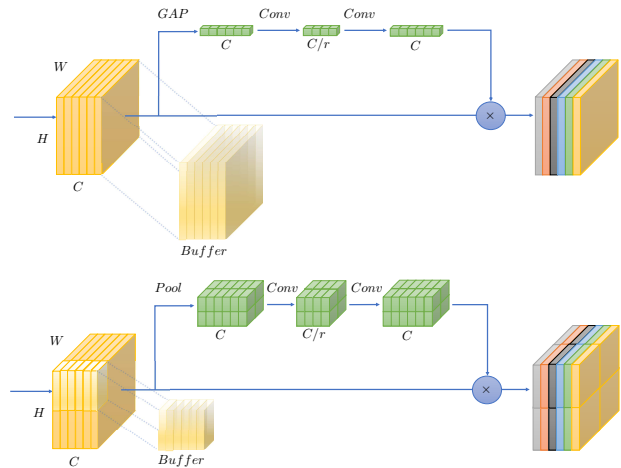}
\caption{On top, the original squeeze-and-excite (SE) block with global average pool (GAP) and on the bottom our proposed tiled squeeze-and-excite (TSE) with a smaller pooling kernel. Smaller kernel uses smaller spatial context and induces a smaller buffer for the element-wise multiplication in AI accelerators with dataflow design.}
\label{op_struecture_fig}
\end{figure}

Channel attention an important building-block in many modern deep learning architectures. An efficient and popular method for channel attention is squeeze-and-excite (SE) \cite{se_paper}. Its structure is comprised of two different steps. The first one being the "squeeze" operation, typically performed by global average pooling (GAP). The objective of this step is to generate a channel descriptor that encodes the global spatial context of the channel. The second step is an "excitation" operation, which produces a collection of per-channel scaling factors to re-calibrate the output tensor. The SE block can be plugged into any common CNN architecture to obtain accuracy improvement with negligible additional compute and parameters; SE blocks have been successfully applied to a variety of computer vision tasks, including classification, detection, and segmentation \cite{efnet_paper,efdet_paper,centermask_paper,dual_paper}. Due to it's parsimonious use of compute resources, it is also heavily used by architectures that are aimed for the mobile compute regime \cite{mnas_paper, ofa_paper, mbv3_paper, shufv2_paper}.

Although SE is highly efficient in terms of compute, its reliance on GAP comes at a cost and might prevent running the network efficiently on some AI accelerators. This is because, unlike GPUs/CPUs, many AI accelerators have a highly-efficient dataflow pipeline design which leverage data reuse. In such accelerators, a minimal amount of information is buffered to produce the output \cite{eyeriss_paper1,eyeriss_paper2,ai_survey_paper}. However, The GAP in SE requires storing the entire input feature map to be multiplied by the output of the "excite" operation (Figure \ref{op_struecture_fig}). Therefore, unlike other parts of the network (\eg convolutions which allows streaming), when doing channel attention via SE the pipeline must be stopped. Since channel attention is used multiple times in the network (\eg in each residual block in SENet) it incurs a noticeable amount of latency. Using a smaller kernel for pooling in the squeeze operation will therefore reduce the required memory of the operation, and enable a streamlined solution for deep learning deployment on AI accelerators.

\par Motivated by the above, in this work, we seek to analyze the minimal amount of spatial context needed for effective channel attention. Based on the SE block, we carefully design a new family of operations, named tiled squeeze-and-excite (TSE), that shares a similar structure with the original operation used in SENet, but works on tiles with limited spatial extent. We show that a limited amount of spatial context is enough for the network to learn meaningful attention factors for each channel and gain the same accuracy achieved by the original operation. This insight is crucial to the understanding of the operation because it suggests that the global spatial context introduced by the GAP in SENet is not needed.

%\par In this work, we attack this problem by analyzing what is the minimal amount of spatial context required for channel attention and show that the same accuracy gains can be obtained with more efficient and streamlined solution for deep learning deployment on AI accelerators. This enables highly efficient usage of channel attention on different hardware architectures such that work in row or patch stationary \cite{ai_survey_paper} and avoid the large memory requirement of the original operation. 

%\par Here, we empirically analyze the amount of spatial context needed for effective channel attention. Based on the SE block, we carefully design a new family of operations, namely tiled squeeze-and-excite (TSE), that shares similar structure to the original operation used in SENet, but works exclusively on different spatial tiles. We show that a limited amount of spatial context is enough for the network to learn meaningful attention factors for each channel and gain the same accuracy achieved by the original operation. This insight is crucial to the understanding of the operation because it suggest that the global spatial context introduced by the GAP in SENet is not needed.

\par
\begin{figure}
\centering
\includegraphics[width=0.47\textwidth]{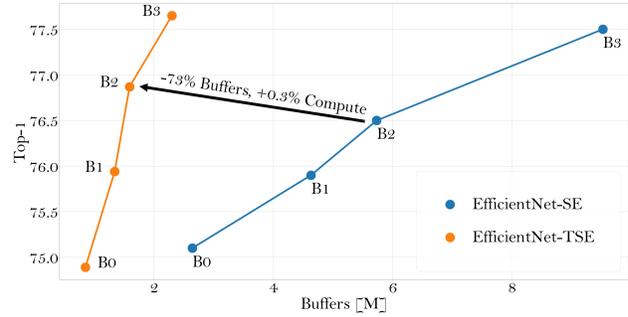}
\caption{Top-1 accuracy and pipeline memory buffer comparison on ImageNet-1K between different models of EfficientNet \cite{efnet_paper} using SE and our proposed TSE.}
\label{models_fig}
\end{figure}

 The proposed TSE solution shares the same "excitation" structure with traditional SE and only differs in the structure of the "squeeze". Sharing the excitation part of the op permits the usage of the original weights of the network (without adding new parameters) and enables fast adaption to the new structure acting as a drop-in replacement. While using TSE incurs more computations (since we need to repeat the excitation processing for each tile), the amount is negligible as illustrated in Figure \ref{models_fig}, which demonstrates that for EfficientNet-B2 \cite{efnet_paper} TSE only adds 0.3\% to the network floating point operations (FLOPs) but uses 73\% less pipeline buffers.

\par Our main contributions can be summarized as follows:
\begin{itemize}
\item We study the local and global spatial context trade-off and show it is sufficient to use local spatial context for effective channel attention.
\item We design a new family of channel attention operators, named tiled squeeze-and-excite (TSE), that can run efficiently on common AI accelerators with data flow design.
\item To demonstrate our solution, we choose a specific variant that is optimized for row-stationary data flow accelerators \cite{ai_survey_paper} and show it has comparable results across different network architectures in image classification, object detection and segmentation.
\end{itemize}

%-------------------------------------------------------------------------
%----------------------- Related Work ------------------------------------
%-------------------------------------------------------------------------

\section{Related Work}

Squeeze-and-excitation networks \cite{se_paper} are a very popular architectural building block with a simple mechanism for channel attention. To improve the accuracy of the original operation, several works used spatial-attention either in parallel or in series \cite{cbam_paper,GALA_paper, spat_att_paper,dual_paper}. Recently, TA \cite{ta_paper} proposed to attend to each tensor dimension separately and fuse the results. Another line of inquiry is better ways to aggregate the global spatial context \cite{ge_paper,gsop_paper, gcnet_paper,strip_pool_paper,fcanet_paper}. ECANet \cite{eca_paper} proposed a more efficient excitation operator using 1D convolutions. Nevertheless, no SE variant has surpassed SE in popularity and use. SE block are an essential building block of many neural architecture searches (NAS) for mobile and edge devices \cite{ofa_paper, mbv3_paper, regnet_paper, efnet_paper}. Specifically, the analysis done in \cite{regnet_paper} showed that the design space which includes the SE block has higher accuracy models than a comparable design space without it.

\par Another line of research studied the spatial context required for attention. Non-local network \cite{nl_paper} showed that learning long-range dependencies is important for tasks which include a temporal dimension such as video classification. They also showed that long range dependencies are useful for object detection. Further analysis by \cite{gcnet_paper} showed that, in practice, the attended context is position independent and therefore the block can be simplified. Both of these works use global spatial-context. Concurrent to our work, coordinate attention \cite{coord_att_paper} proposed to look at limited spatial context composed of both vertical and horizontal strips. Similarly to us, they limit the amount of spatial context used to strips of one row or column and show that they can match the performance of MobileNetV2 \cite{mbv2_paper} with SE. However, the focus of their work is on improving the accuracy of SE and so they provide no further analysis or experiments on that front.

\par The work most closely related to ours is the Gather-Excite (GE) framework \cite{ge_paper}. The framework introduced in the paper is very suitable for studying the effects of spatial-context on channel attention and with slight modifications to their formulation our block of TSE can be seen as an instantiation of GE. Nonetheless, the focus of their work is how to better encode global spatial-context to get maximal accuracy.

\par Different from all the above, our work is focused on exploring the amount of spatial context required for effective channel-attention and not on improving the accuracy of the SE operation. Compared with other works, our TSE block relies solely on local spatial context.
% They only briefly compare the performance of local spatial-context to global-context and reach the conclusion that global context gives slightly better performance. We reach a different conclusions and note that they experimented only with patch-tiles which require a larger extent than the one used in their experiments.

%-------------------------------------------------------------------------
%----------------------- Method ------------------------------------------
%-------------------------------------------------------------------------

\section{Tiled Squeeze-and-Excite} \label{tiled_SE_sec}

In this section, we present tiled squeeze-and-excite, a framework for exploring the amount of spatial context required for effective channel attention.  Tiled squeeze-and-excite is a modification of the SE block \cite{se_paper} which preservers the numbers of parameters but enables to vary the spatial-context used. %We show that spatial-context is not 'free' and that there is an inherent compute/buffering trade-off which can be exploited by different types of AI accelerators hardware architectures and therefore 
Our goal is to find the minimum spatial context required to match the accuracy achieved with global context. 

\subsection{Tiled Squeeze} \label{tiled_squeeze_sec}

We begin with a review of the original squeeze-and-excite block. Given a tensor of dimensions $T\in\mathbb{R}^{H \times W \times C}$, the squeeze-and-excite operator re-scales each channel by a scalar in the range [0,1]. The "squeeze" operation is parameter-free and encodes the global spatial context of each channel into a single descriptor by means of global average pooling (GAP). It's output is a tensor of dimensions $Z\in\mathbb{R}^{1 \times 1 \times C}$. The "excite" operation performs channel-attention using a two-layer fully-connected feed-forward network described by $\sigma(W2\cdot ReLU(W_1Z))$ where $W_1 \in \mathbb{R}^{C\times C/r}$, $W_2\in\mathbb{R}^{C/r\times C}$ and $\sigma$ denotes the sigmoid activation function. The output of the excite operation is a vector of length $C$ which is than used to scale the channels of the input tensor via element-wise multiplication. Note that the SE block has a built-in separation between spatial-context encoding (squeeze) and channel-attention (excite). Since we are interested in studying the effect of spatial-context encoding without changing the channel-attention mechanism, we keep the excite operation as is. To keep alignment with the original SE block we also wish to avoid adding parameters to the squeeze operation. Thus, what we want is a parameter-free squeeze block whose spatial-context can be varied. 

\begin{figure}
%\centering
\begin{subfigure}[t]{0.5\columnwidth}
    \setlength{\medmuskip}{0mu}
    \psfrag{A}[cc][cc][0.8]{$GAP$}
    \psfrag{B}[cc][cc][0.8]{$FC\,(C\rightarrow C/r)$}
    \psfrag{C}[cc][cc][0.8]{$ReLU$}
    \psfrag{D}[cc][cc][0.8]{$FC\,(C/r\rightarrow C)$}
    \psfrag{E}[cc][cc][0.8]{$Sigmoid$}
    \psfrag{x}[cl][cc][0.6]{$W \times H \times C $}
    \psfrag{y}[cl][cc][0.6]{$1 \times 1 \times C$}
    \psfrag{z}[cl][cc][0.6]{$1 \times 1 \times C/r$}
    \psfrag{w}[cl][cc][0.6]{$1 \times 1 \times C$}
    \includegraphics[width=0.95\linewidth]{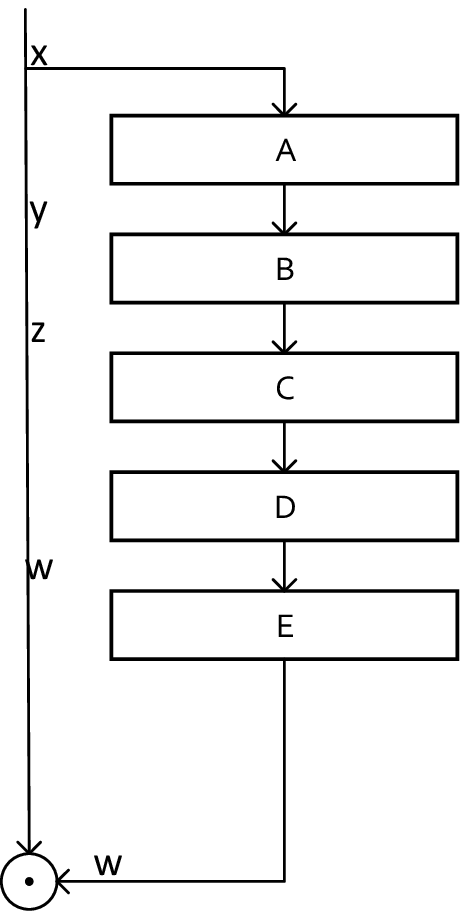}\hfill
    \caption{SE block}
\end{subfigure}%
\begin{subfigure}[t]{0.5\columnwidth}
\setlength{\medmuskip}{0mu}
    \psfrag{A}[cc][cc][0.8]{$AvgPool\, (h,w/h,w)$}
    \psfrag{B}[cc][cc][0.8]{$Conv1\times1\,(C\rightarrow C/r)$}
    \psfrag{C}[cc][cc][0.8]{$ReLU$}
    \psfrag{D}[cc][cc][0.8]{$Conv1\times1\,(C/r\rightarrow C)$}
    \psfrag{E}[cc][cc][0.8]{$Sigmoid$}
    \psfrag{F}[cc][cc][0.8]{$Interpolation$}
    \psfrag{x}[cl][cc][0.6]{$W \times H \times C $}
    \psfrag{y}[cl][cc][0.6]{$\lceil H/h \rceil \times \lceil W/w \rceil \times C$}
    \psfrag{z}[cl][cc][0.6]{$\lceil H/h \rceil \times \lceil W/w \rceil \times C/r$}
    \psfrag{w}[cl][cc][0.6]{$\lceil H/h \rceil \times \lceil W/w \rceil \times C$}
    \psfrag{t}[cl][cc][0.6]{$W \times H \times C$}
    \hfill\includegraphics[width=0.95\linewidth]{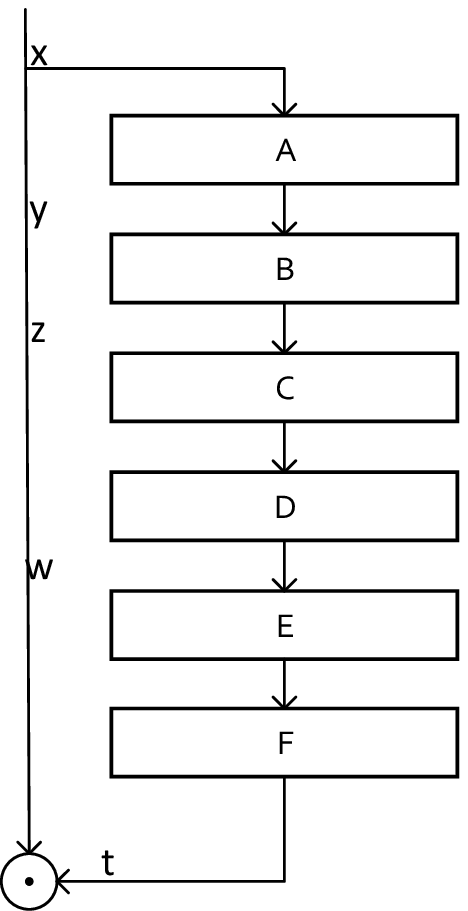}
    \caption{Tiled-SE block}
\end{subfigure}
\centering
\caption{Block diagram of the proposed block. (a) The original SE block. (b) the Tiled-SE block. Tiled-SE uses average-pooling with limited extent in-place of the global pooling. The stride of the pool is the same as the kernel dimension so tiles are non-overlapping. Before scaling, the tiles are broadcast back to the dimensions of the input tensor using nearest-neighbour interpolation.}
\label{op_block_diagram_fig}
\end{figure}
\par Our proposed concept is as follows: we spatially partition the input tensor to non-overlapping tiles of equal size. The channels of each tile are then re-scaled by SE block which is shared for all tiles. The re-scaled tiles are then stitched back together to get the output tensor. We term this operation - tiled squeeze-and-excite (TSE). Since the channel-attention relies on aggregated spatial context, we expect that as we increase the tile size, the attention mechanism will become more effective and the performance of the network will improve. Tiled-SE is illustrated in Figure \ref{op_struecture_fig}.  The implementation of TSE is simple: To produce tiles, we replace the GAP of the SE block with average pooling with both kernel and stride matching the size of the tile - $AvgPool2D((h,w))$. The number of tiles produced is $N=\lceil \dfrac{H}{h}\times \dfrac{W}{w}\rceil$. The fully-connected layers are changed to 1x1 convolutions. The output of the sigmoid is resized to the dimensions of the input tensor using nearest-neighbour interpolation. A block-diagram of the implementation is shown in Figure \ref{op_block_diagram_fig} and the matching PyTorch code is given in the supplementary material.
\par We note that it is possible to change many of the design selections made in TSE (see supplementary material for different configuration). Here we opt for the simplest configuration without any bells-and-whistles due to two considerations: (1) we want to isolate the effect of using local-context from any other changes to the architecture and (2) we want to stay compatible with the original SE block. In later sections, we show that the interchangeability of SE and TSE means we can interpret TSE as an estimator of SE (or as a noisy approximation of SE). In our experiments we show that it allows us to plug-in TSE into models trained with SE without re-training.

\subsection{TSE Instantiations} \label{tile_instance_sec}
Switching from global spatial-context to tiles introduces a new hyper-parameter of tile size. In this section, we study how different tile sizes affect the performance of TSE. We choose to investigate two types of tiling strategies:
\begin{enumerate}
  \item \textbf{Strip-tiling} in which tiles are composed of $k$ strips of either rows or columns. $k$ is constant across the network. Note that when using strip-tiling the size of the tile (as measured by amount of elements in the tile) decreases as the spatial dimension of the network is decreased but the ratio of tile-size to tensor-size increases since $k$ becomes larger with respect to spatial dimensions of the tensor. We denote by $TSE_{k \times W}, TSE_{H \times k}$ row and columns tiles respectively. For simplicity, when discussing strip-tiling in the text we will implicitly mean row-tiling unless explicitly stated otherwise.
  \item \textbf{Patch-tiling} in which tiles are composed of fixed-size $k\times k$ patches. $k$ is constant across the network. Note that while the size of the tile remains constant the ratio of tile-size to tensor-size increases as the spatial dimension of the network decreased. We denote by $TSE_{k \times k}$ patch-tiles.
\end{enumerate}
 For a given input-tensor a change in tile-size affects three metrics: accuracy, compute and the pipeline buffering. We thoroughly discuss the effect on accuracy as we vary the tile size in Section \ref{global_vs_local_sec}. Before that, we briefly review the implications for compute and buffering herein.
 
 \begin{figure*}
\centering
\psfrag{A}[cc][cc][1.0]{SE$_{2,3}$}
\psfrag{B}[cc][cc][1.0]{SE$_{3,4}$}
\includegraphics[scale=1.0]{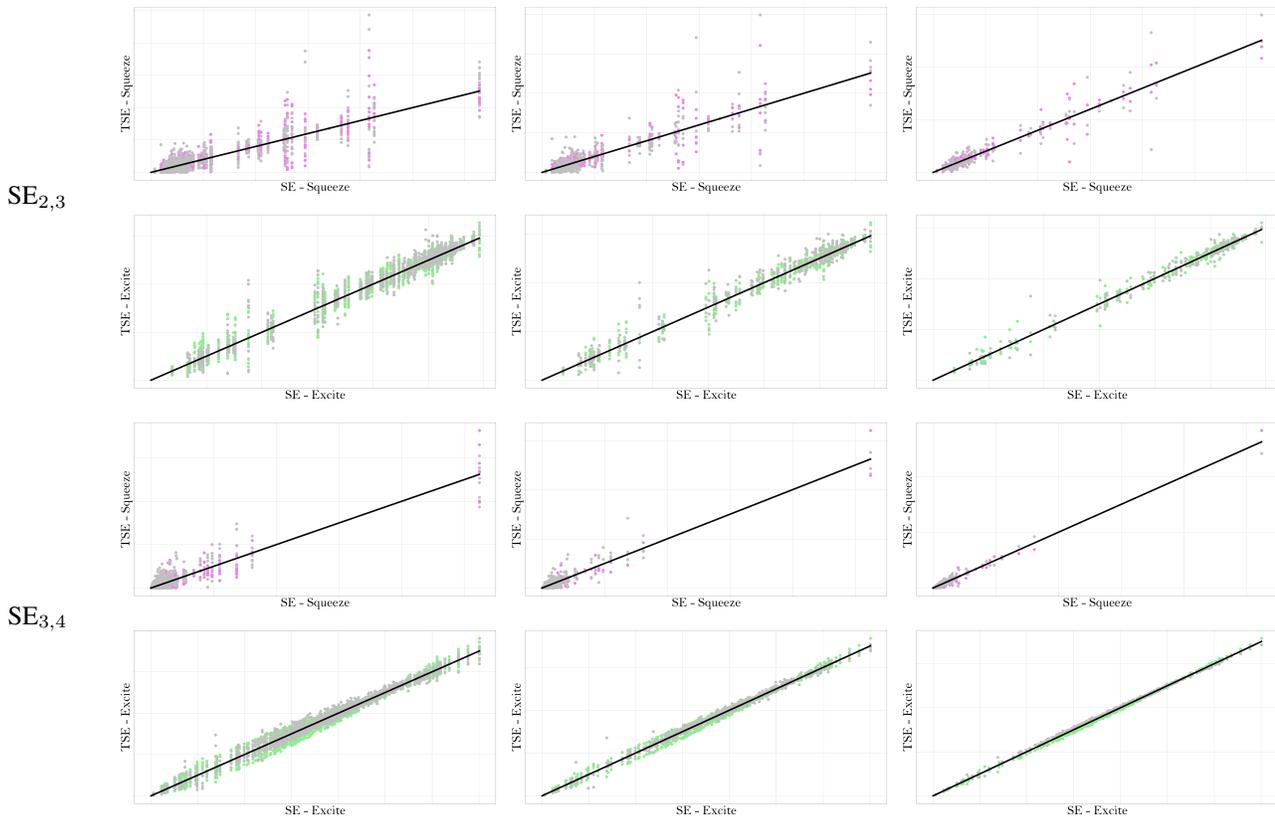}
\caption{Correlation map induced by the squeeze operation (purple) and excite operation (green) at different depths in RegNetY-800MF \cite{regnet_paper} on ImageNet-1K. On top, stage 2 block 3 and on the bottom stage 3 block 4. The colored dots represents the 50\% of the centred tiles. This figure is best viewed in color.}
\label{correlation_fig}
\end{figure*}
 
\par\textbf{Compute.} We denote the additional compute introduced by an individual SE block as $F$. The compute of the corresponding TSE block is therefore $N\cdot F$ where $N$ is the number of tiles used by TSE. Generally speaking, the additional compute introduced by each SE block is small and for strip-tiling the number of tiles $N$ is small (\eg in the order of few millions FLOPs in EfficientNet-B3) so the overall increase in compute is negligible. However, since the number of tiles scales with input resolution (linearly for strip-tiling and quadratically for patch-tiling), the additional compute may become non-negligible at high-resolutions. From a compute perspective we would rather select large tiles but for most networks the modest additional compute is offset by a significant rise in network accuracy. Therefore, we treat compute as a secondary consideration.

\par\textbf{Buffering.} As discussed in Section \ref{intro_sec}, one of the motivations for working with limited spatial-context is to minimize the required pipeline buffering in dataflow architectures \cite{ai_survey_paper}. Ignoring implementation details, the minimum buffering required for TSE is $h\times w \times C$, where $h,w$ are the spatial dimensions of the tile. For strip-pooling the amount of required buffering is $k \times W \times C$ and it is linear in both $W$ and $k$. As the input resolution increases so does the amount of required buffering. For patch-tiling the required buffering is $k \times k \times C$ which is quadratic in $k$ but independent of the spatial dimensions of the tensor. The amount of actual buffering required depends on the HW implementation. For example, row-stationary architectures \cite{eyeriss_paper1, eyeriss_paper2} have pipeline-buffering granularity that is $\propto W$ and therefore can benefit most from strip-tiling. From a buffering perspective we would rather select small tiles.

%The amount of actual buffering required depends both on the the specifics of the hardware use for the deployment and software kernel implementation. For example, row-stationary architectures \cite{eyeriss_paper1, eyeriss_paper2} have pipeline-buffering granularity that is $\propto W$ and therefore can benefit most from strip-tiling. From a buffering perspective we would rather select small tiles.

\subsection{Local Spatial Context for Channel Attention} \label{global_vs_local_sec}
As mentioned in section \ref{tiled_squeeze_sec}, TSE is compatible to SE and can therefore be plugged-in instead of SE in a network after it was trained with the SE block. If the local spatial context within each tile is a good estimation of the global spatial context than such a replacement should result in minimum performance degradation. Thus, we treat each tile's mean as an estimation of the global mean. If we assume that (1) activations are distributed homogeneously across the spatial dimension and (2) that the number of sampling points in each tile is large compared to the variance of the activation, than we are guaranteed (in the mean)  to obtain good performance when replacing SE by TSE. More formally, we denote the GAP of channel $i$ as $G_i$ and the mean of tile $j$ of the same channel as $T_i^j=G_i + \delta_i^j$ where the $\delta$ denotes the difference in the mean of the tile compared to the GAP. If we denote the number of points in the tile as $n$ and the variance of the activations in the channel as $\sigma_i^2$ then $\delta_i \propto\sqrt{\sigma_i/n}$. %We denote by $\boldsymbol{g}$, $\boldsymbol{\delta^j}$ the channel descriptor vectors for the GAP and tile $j$ respectively.
Then the scale vector $S^j_i$ estimated by tile $j$ is given by:
\begin{equation} \label{TSE_noise_eq}
S^j_i =\sigma(W2\cdot ReLU(W_1\cdot (G_i+\delta^j_i))
\end{equation}
Equation \ref{TSE_noise_eq} shows that we can treat TSE as a noisy approximation of SE.

\par Previous works also showed that SE has larger contribution in deeper layers \cite{se_paper, ge_paper}. Therefore, to match the performance of SE, we would want tiles that become progressively larger. This is exactly what TSE achieves: the tile-to-tensor size ratio is increased for deeper layers, making TSE a progressively better estimator. We verify the above analysis by the following experiment: we take a network trained with SE and we replace one of the layers with strip-pooling TSE without retraining. Then we measure the correlation between the GAP of the layers and the means of the tiles. We test the correlation both after the squeeze operation and after the excite operation. We do the experiment with different values of $k$ and we select shallow and deep layers. The results are shown in Figure \ref{correlation_fig}. As expected, the correlation improves as we increase the tile size.

\section{Experiments}
\par In this section we first study the impact of different tile sizes in TSE and empirically show the viability of using smaller spatial context to learn meaningful channel attention factors. Next, we evaluate the performance of TSE when used as a replacement in networks pre-trained with SE. These experiments confirms that TSE can be used in existing SE networks without re-training or with a short fine-tuning step. Finally, we report the results of TSE on variety of models in image classification, object detection and semantic segmentation. We show that TSE generalizes across different models, different tasks and a wide scale of input resolutions.

%In this experiment, we adopt RegNetY-800MF model \cite{regnet_paper} and only modify the pooling kernel of its TSE block throughout the network.

\subsection{Implementation Details}\label{imp_sec}
We evaluate TSE on ImageNet-1K \cite{imagenet_paper} for image classification, MS COCO \cite{coco_paper} for object detection and Cityscapes \cite{cityscapes_paper} for semantic segmentation. To make the comparison to SE baseline meaningful, we reproduce all SE results using the same framework as we use for TSE. The image classification models were implemented using the pycls\footnote{https://github.com/facebookresearch/pycls} toolkit. Each model was trained with 8 V100 GPUs for 100 epochs using stochastic gradient descent (SGD), momentum of 0.9 and weight decay of 5e-5. The base learning rate was set to 0.8 for the RegNet \cite{regnet_paper} models and to 0.4 for the EfficientNet \cite{efnet_paper} models. We follow the cosine learning policy for updating the learning rate during training \cite{sgdr_paper}. Our results were obtained with a short training schedule and without enhancements.

\par The object detection models were implemented in EfficientDet-PyTorch\footnote{https://github.com/rwightman/efficientdet-pytorch} toolkit and we optimized the models for 300 epochs using an SGD optimizer, momentum of 0.9 and weight decay of 4e-5. The base learning rate was set to 0.08 and we updated it according to the cosine decay method. For augmentation, we only used random flip and resize without special enhancement.

\par The semantic segmentation models were implemented using the MMSegmenation\footnote{https://github.com/open-mmlab/mmsegmentation} toolkit. We optimized the models using SGD for 160k iterations with base learning rate of 1e-2, momentum of 0.9 and weight decay of 5e-4.
 
\subsection{Channel Attention with Local Spatial-Context}
Here, we examine how the accuracy of the network changes as we vary the tile size in TSE. All experiments are done with the RegNetY-800MF \cite{regnet_paper} architecture. We make several empirical claims:
\par\textbf{Local-context is sufficient}. We train the model using three different configurations: strip-tiling of rows (TSE$_{k\times W}$), strip-tiling of columns (TSE$_{H\times k}$) and patch-tiling (TSE$_{k \times k}$). For strip-tiling we vary $k$ from 1 to 9 and for patch-tiling we vary $k$ from 1 to 13 since patch-tiles are smaller than strip-tiles. We compare the results to a baseline when the model is trained with and without SE. The full results are given in Table \ref{context_tab}. We see that for all tiling strategies, we are able to match the performance of SE using only a portion of the global spatial context. Thus, we show that effective channel attention does not require global context. For strip-pooling a value of $k=7$ is sufficient to match the accuracy of SE while for patch-pooling $k=13$ is needed in order to be on par. We note that the spatial dimension of the last feature map in our model is $7\times 7$ so for $k=7$ SE and TSE converge.

\begin{table}
\setlength\tabcolsep{3pt}
\scriptsize
\centering
\begin{tabular}{@{}lllll@{}}
\toprule
Method & Params & MFLOPs & Buffer & Top-1 \\
\midrule
Vanilla & 5.4M & 796.35 & N/A & 75.07 \\
SE & 6.2M & 797.18 & 1.07M & 76.30 \\
\midrule
TSE$_{7\times W-upper}$ & 6.2M & 797.18 & 1.07M & 75.88 \\
TSE$_{7\times W-middle}$ & 6.2M & 797.18 & 1.07M & 76.25 \\
\midrule
TSE$_{9 \times W}$ & 6.2M & 797.68 & 0.52M & 76.32 \\
TSE$_{7 \times W}$ & 6.2M & 797.88 & 0.42M & 76.29 \\
TSE$_{5 \times W}$ & 6.2M & 798.49 & 0.30M & 75.98 \\
TSE$_{3 \times W}$ & 6.2M & 799.92 & 0.18M & 76.00 \\
TSE$_{1 \times W}$ & 6.2M & 807.03 & 0.06M & 75.79 \\
\midrule
TSE$_{H \times 9}$ & 6.2M & 797.68 & 0.52M & 76.49 \\
TSE$_{H \times 7}$ & 6.2M & 797.88 & 0.42M & 76.42 \\
TSE$_{H \times 5}$ & 6.2M & 798.49 & 0.30M & 76.15 \\
TSE$_{H \times 3}$ & 6.2M & 799.92 & 0.18M & 75.74 \\
TSE$_{H \times 1}$ & 6.2M & 807.03 & 0.06M & 76.07 \\
\midrule
TSE$_{13 \times 13}$ & 6.2M & 797.58 & 0.58M & 76.34 \\
TSE$_{11 \times 11}$ & 6.2M & 797.83 & 0.43M & 76.15 \\
TSE$_{9 \times 9}$ & 6.2M & 798.03 & 0.36M & 76.02 \\
TSE$_{7 \times 7}$ & 6.2M & 799.11 & 0.22M & 76.06 \\
TSE$_{5 \times 5}$ & 6.2M & 801.76 & 0.11M & 75.80 \\
TSE$_{3 \times 3}$ & 6.2M & 811.35 & 0.04M & 75.44 \\
TSE$_{1 \times 1}$ & 6.2M & 931.23 & N/A & 75.54 \\
\bottomrule
\end{tabular}
\caption{Comparison of different tiles in TSE on RegNetY-800MF \cite{regnet_paper} network with ImageNet-1K. TSE$_{h \times w}$ stands for tile size $h \times w$. The buffer column indicates the minimum amount of pipeline buffering required for the op throughout the network as detailed in Section \ref{tile_instance_sec}.} 
\label{context_tab}
\end{table}

\par\textbf{Not all locations are equal}. To test whether performance is determined by tile size or tile location, we train two additional variants: TSE$_{7 \times W-upper}$ and TSE$_{7 \times W-middle}$. Both variants use a \textit{single} row strip tile. TSE$_{7 \times W-upper}$ always uses the upper seven rows while TSE$_{7 \times W-middle}$ always uses a row strip centered around the middle row of the tensor. If tile size would be the only factor determining performance we would expect to see both performing on par. Instead, we see that TSE$_{7 \times W-middle}$ far outperforms TSE$_{7 \times W-upper}$ and, moreover, it performs on par with the original SE and TSE$_{7 \times W}$ suggesting that the center of image holds most of the 'interesting' spatial context. We note, that in ImageNet most images contain an object located in the center of the image and that the above result depend on the dataset and task.

\begin{figure}
\centering
\begin{subfigure}[b]{0.475\textwidth}
    \psfrag{A}[cc][cc][0.5]{TSE$_{k \times W}$}
    \psfrag{B}[cc][cc][0.5]{TSE$_{H \times k}$}
    \psfrag{C}[cc][cc][0.5]{TSE$_{k \times k}$}
    \centering
    \includegraphics[scale=0.125]{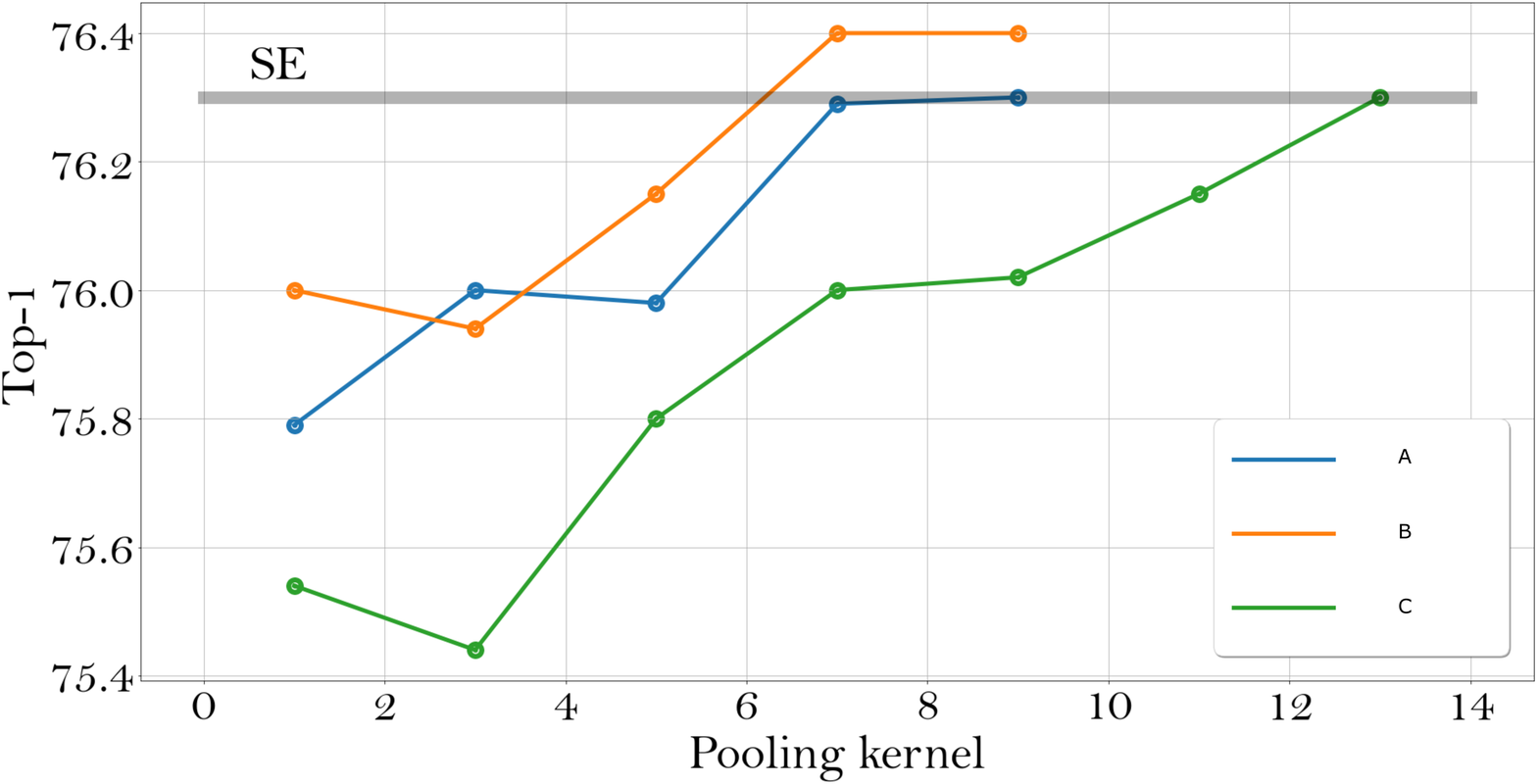}
    \caption[With Training]{{\small With Training}}
    \label{trend_train_fig}
\end{subfigure}
\vskip\baselineskip
\begin{subfigure}[b]{0.475\textwidth}
    \centering
    \psfrag{A}[cc][cc][0.5]{TSE$_{k \times W}$}
    \psfrag{B}[cc][cc][0.5]{TSE$_{H \times k}$}
    \psfrag{C}[cc][cc][0.5]{TSE$_{k \times k}$}
    \centering
    \includegraphics[scale=0.125]{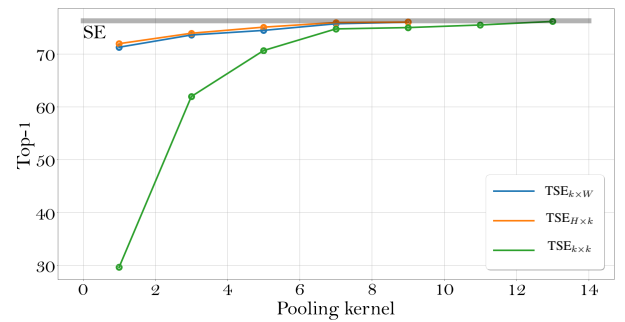}
        \caption[Without Training]{{\small Without Training}}
    \label{trend_notrain_fig}
\end{subfigure}
\caption{Pooling tile trends in TSE on ImageNet-1K. In TSE$_{k \times W}$ and TSE$_{H \times k}$ we change the pooling kernel by increasing one dimension and fixing the other and in TSE$_{k \times k}$ we change both spatial dimensions.}
\label{pool_fig}
\end{figure}

\par\textbf{The learned channel-attention is a function of tile size}: In section \ref{global_vs_local_sec} it was noted that when replacing SE with TSE without training the performance of the network should degrade only slightly if the tile sizes are large enough. A different way to phrase it is that the channel attention learned with global spatial context will also work for local-context, provided that the local context is large enough. In Figure \ref{pool_fig}, we plot the network accuracy as function of tile size under two scenarios: training the network with TSE from scratch and training the network with SE and post-training replacing the SE with TSE. We see two interesting phenomena. For both strip and patch tiles of size $k=7$ the local-context are good enough estimators of the global spatial context. On the other hand for smaller values of $k$ we see a significant degradation of the results when trying to apply channel-attention learned from global-context to local-context. However, when training from scratch with TSE we see that some channel-attention can be learned even without any context (e.g. TSE$_{1,1}$) and the accuracy improves over the baseline model trained without any spatial context. This suggests that for small tiles the channel-attention is different that the one learned when global-context is available.
%The activation memory buffer reported in Table \ref{context_tab} is calculated as the minimum amount of pipeline buffering required in order to compute the TSE operation throughout the network. For example, using TSE$_{H,W}$ requires the entire input tensor for the element-wise multiplication in each block and TSE$_{1,W}$ requires only a single row.

% \par In Figure \ref{pool_fig}, we show the trends for increasing the spatial context on three different configuration: strip-tiling of rows (TSE$_W$), strip-tiling of columns (TSE$_H$) and patch-tiling (TSE$_P$). For strip-tiling we vary $k$ from 1 to 9 and for patch-tiling we vary $k$ from 1 to 13 since patch-tiles are smaller than strip-tiles. It is evident that as we increase $k$ (e.g. the size of initial tile) the accuracy improves. For strip-pooling a value of $k=7$ is enough to match the accuracy of SE while for patch-pooling $k=13$ is needed in order to be on par. We note that the spatial dimension of the last feature map in our model is $7\times 7$ so for $k=7$ SE and TSE converge for these layers. The accuracy trends with respect to $k$ are plotted in figure \ref{trend_train_fig} for training the network from scratch and in \ref{trend_notrain_fig} for inference without training. The results evidently show that \textit{local-context is sufficient for effective channel attention}.

In all the following sections we adopt a single TSE variant - TSE$_{7 \times W}$ - and perform all subsequent experiment only with it.

%\subsection{Using Pre-Trained Weights}\lablel{sec_pretrained}
\subsection{Using Pre-Trained Weights}\label{sec_pretrained}
In this section, we evaluate on a wide variety of networks the performance of TSE$_{7 \times W}$ when used as a post-training replacement of SE. We adopt the EfficientNet \cite{efnet_paper} and RegNet \cite{regnet_paper} family of models for evaluation. The results are presented in Table \ref{pretrained_tab}. We see that for most networks the degradation is 0.6\% or below but some network exhibit higher degradation. To get some better insight, we take EfficientNet-B3 and RegNetY-3.2GF, the models with the highest degradation in each family, and give a full breakdown of the degradation in each. For this experiment, we only replace part of the SE blocks with TSE and measure degradation. Results are presented in Table \ref{deg_tab}. We see that degradation is additive and that it mostly arises at later stages of the network. Additionally, in the first stages of the network we can employ TSE blocks with almost zero degradation despite being spatially larger which suggests earlier stages don't require global spatial context. This confirms previous findings \cite{se_paper,ge_paper} that channel attention is more valuable for deeper layers.

\begin{table}
\scriptsize
\centering
\begin{tabular}{@{}lllll@{}}
\toprule
Model & Input & SE & TSE$_{7 \times W}^{aw}$ & TSE$_{7 \times W}^{ft}$ \\
\midrule
RegNetY-200MF & 224$\times$224 & 70.3 & 70.1\tiny{(0.2)} & 70.13\tiny{(0.16)} \\
RegNetY-400MF & 224$\times$224 & 74.1 & 73.7\tiny{(0.6)} & 73.94\tiny{(0.16)} \\
RegNetY-600MF & 224$\times$224 & 75.5 & 75.1\tiny{(0.4)} & 75.23\tiny{(0.27)} \\
RegNetY-800MF & 224$\times$224 & 76.3 & 75.7\tiny{(0.6)} & 75.96\tiny{(0.34)} \\
RegNetY-1.6GF & 224$\times$224 & 77.9 & 77.6\tiny{(0.3)} & 77.60\tiny{(0.30)} \\
RegNetY-3.2GF & 224$\times$224 & 78.9 & 78.2\tiny{(0.7)} & 78.77\tiny{(0.13)} \\
\midrule
EfficientNet-B0 & 224$\times$224 & 75.1 & 74.6\tiny{(0.43)} & 74.96\tiny{(0.14)} \\
EfficientNet-B1 & 240$\times$240 & 75.9 & 75.1\tiny{(0.80)} & 75.72\tiny{(0.18)} \\
EfficientNet-B2 & 260$\times$260 & 76.5 & 75.5\tiny{(0.92)} & 76.20\tiny{(0.30)} \\
EfficientNet-B3 & 300$\times$300 & 77.5 & 76.3\tiny{(1.14)} & 77.14\tiny{(0.36)} \\
\bottomrule
\end{tabular}
\caption{Comparing Top-1 accuracy of TSE networks with pre-trained weights of SE on ImageNet-1K. The results for TSE$_{7 \times W}^{aw}$ are obtained by assigning all the weights from the SE network into the TSE$_{7 \times W}$ (without training) and TSE$_{7 \times W}^{ft}$ is the same network after fine-tuning. Degradation compared to SE baseline is noted in parenthesis. }
\label{pretrained_tab}
\end{table}

\par Next, we wish to see if performance can be regained by doing a short fine-tuning step. Post TSE replacement, we train the networks a further 40 epochs on a subset of 10\% of the ImageNet-1K training data. Results are given in Table \ref{pretrained_tab}. We see that after fine-tuning, all the models converged to the baseline result with a small degradation (less than 0.4\%). Taken together, these experiments validate that TSE can be used as a post-training replacement for SE. A practical implication is that an SE network can be trained once and deployed on different types of hardware.

\begin{table}
\setlength\tabcolsep{3pt}
\scriptsize
\centering
\begin{tabular}{@{}lll@{}}
\toprule
Stage & EfficientNet-B3 & RegNetY-3.2GF \\
\midrule
Baseline & 77.5 & 78.9 \\
\midrule
  Stage-1 & 77.39\tiny{(0.11)} & 78.9\tiny{(0.00)} \\
  Stage-2 & 77.42\tiny{(0.08)} &  78.84\tiny{(0.06)} \\
  Stage-3 & 77.50\tiny{(0.00)} &  78.39\tiny{(0.51)} \\
  Stage-4 & 77.39\tiny{(0.11)} &  78.9\tiny{(0.00)} \\
  Stage-5 & 77.48\tiny{(0.02)} & - \\
  Stage-6 & 77.20\tiny{(0.30)} & - \\
  Stage-7 & 77.01\tiny{(0.49)} & - \\
\midrule
 TSE$^{aw}$ & 76.36 & 78.2 \\
\bottomrule
\end{tabular}
\caption{Top-1 degradation breakdown of assigning SE weights into a TSE in different networks on ImageNet-1K. In each stage we replace all the SE blocks with TSE and measure the degradation. The stage terminology is takes from the respective family architectures.}
\label{deg_tab}
\end{table}

\iffalse
\subsection{Comparison with Other Methods}
In this section, we evaluate our method against other channel attention methods that we have tried.

\begin{table}
\setlength\tabcolsep{3pt}
\scriptsize
\centering
\begin{tabular}{@{}ll@{}}
\toprule
Model                                  & Top-1 \\
\midrule
RegNetY-800MF-SE                          & 76.30 \\
RegNetY-800MF-no-SE                    & 75.07 \\
\textbf{RegNetY-800MF-TSE}                    & \textbf{76.29} \\
RegNetY-800MF-7-upper                  & 75.88 \\
RegNetY-800MF-7-middle                 & 76.25 \\
RegNetY-800MF-FcaNet                   & 74.85 \\
RegNetY-800MF-ECANet                  & 75.55 \\
$\theta^-$ \cite{ge_paper}                             & 74.98 \\
\bottomrule
\end{tabular}
\caption{Ablation study for different channel attention methods. The baseline is our experiment of RegNetY-800MF-SE and it is compared to removing the SE blocks (no-SE), using context of only the seven upper rows in each SE block (7-upper), the middle seven rows (7-middle) and the implementation of three works: Eca-Net \cite{eca_paper}, FcaNet \cite{dct_paper} and $\theta^-$ from \cite{ge_paper}.}
\label{fix_context_tab}
\end{table}
\fi

\subsection{Training with TSE}
In the previous section we validated the performance of TSE when used as post-training replacement for SE. In the following sections we evaluate the performance of the TSE block when the network is trained from scratch with TSE.

\subsubsection{Classification}
We perform ImageNet-1K classification experiments to evaluate the TSE block compared to SE. Specifically, we follow the same protocol as specified in Section \ref{imp_sec} and employ TSE$_{7 \times W}$ to verify the accuracy gain is maintained across different architectures. Table \ref{classification_tab} summarizes the experimental results. For all networks examined, TSE has comparable accuracy with models trained with SE. With respect to the results in section \ref{sec_pretrained} we note that training from scratch with TSE gives slightly better results than fine-tuning on networks pre-trained on SE.

\begin{table}
\setlength\tabcolsep{3pt}
\scriptsize
\centering
\begin{tabular}{@{}lllllll@{}}
\toprule
Model & \multicolumn{3}{c}{SE} & \multicolumn{3}{c}{TSE$_{7 \times W}$} \\
& Top-1 & Buffer & GFLOPs & Top-1 & Buffer & GFLOPs \\
\midrule
RegNetY-200MF & 70.3 & 0.38M & 0.2 & 70.53 & 0.20M & 0.2 \\
RegNetY-400MF & 74.1 & 0.76M & 0.4 & 73.87 & 0.33M & 0.4 \\
RegNetY-600MF & 75.5 & 0.88M & 0.6 & 75.43 & 0.37M & 0.6 \\
RegNetY-800MF & 76.3 & 1.07M & 0.8 & 76.29 & 0.42M & 0.8\\
RegNetY-1.6GF & 77.9 & 2.07M & 1.6 & 77.87 & 0.82M & 1.6\\
RegNetY-3.2GF & 78.9 & 2.84M & 3.2 & 78.77 & 1.07M & 3.2 \\
\midrule
EfficientNet-B0 & 75.1 & 2.64M & 0.4 & 74.89 & 0.85M & 0.4\\
EfficientNet-B1 & 75.9 & 4.63M & 0.7 & 75.94 & 1.34M & 0.7\\
EfficientNet-B2 & 76.5 & 5.73M & 1.0 & 76.87 & 1.59M & 1.0\\
EfficientNet-B3 & 77.5 & 9.52M & 1.8 & 77.65 & 2.30M & 1.8\\
\bottomrule
\end{tabular}
\caption{Comparison of Top-1 accuracy results on ImageNet-1K for different EfficientNet \cite{efnet_paper} and RegNetY \cite{regnet_paper} models .}
\label{classification_tab}
\end{table}

\subsubsection{Object Detection}
We evaluate TSE on object detection trained on COCO 2017 \cite{coco_paper}. We employ EfficientDet \cite{efdet_paper} models which are a strong baseline for object detection with extensive usage of SE blocks. Table \ref{effdet_tab} shows that EfficientDet-TSE has comparable accuracy to SE. We make two important observations. Up-until now, we showed that TSE is comparable to SE at low resolutions only, and since spatial-context is intimately related to input resolution it is not given that TSE's performance would scale with resolution. Second, object-detection is a much more spatially-sensitive task compared with classification. Thus, we see that our previous conclusion about local spatial-context being sufficient generalize broadly. We also note that at higher resolutions the cost of pipeline-buffering becomes much more prohibitive and here TSE requires $\times$10 less pipeline buffering compared to SE.

\begin{table}
\scriptsize
\centering
\begin{tabular}{@{}lllll@{}}
\toprule
 & & EfficientDet-D0 & EfficientDet-D1 & EfficientDet-D2 \\
\midrule
SE & mAP & 33.8 & 39.0 & 42.3 \\
   & Buffer & 13.8M & 33.6M & 50.8 \\
   & GFLOPs & 2.5 & 6.1 & 11 \\
\midrule
TSE$_{7 \times W}$ & mAP & 33.9 & 39.6 & 42.3 \\
   & Buffer & 1.9M & 3.6M & 4.7M \\
   & GFLOPs & 2.5 & 6.1 & 11 \\
\midrule
TSE$_{7 \times W}^{aw}$ & mAP & 33.0 & 38.0 & 41.0 \\
   & Buffer & 1.9M & 3.6M & 4.7M \\
   & GFLOPs & 2.5 & 6.1 & 11 \\
\bottomrule
\end{tabular}
\caption{Comparison of mAP accuracy results on MS COCO-2017 validation set for different EfficientDet models \cite{efdet_paper}.}
\label{effdet_tab}
\end{table}

%\begin{table}
%\setlength\tabcolsep{2pt}
%\scriptsize
%\centering
%\begin{tabular}{@{}llllllll@{}}
%\toprule
%Model & Input & \multicolumn{3}{c}{SE} & \multicolumn{3}{c}{TSE} \\
%& & mAP & Buffer & GFLOPs & mAP & Buffer & GFLOPs \\
%\midrule
%EfficientDet-D0 & 512$\times$512 & 33.8 & 13.8M & 2.5 & 33.9 & 1.9M & 2.5 \\
%EfficientDet-D1 & 640$\times$640 & 39.0 & 33.6M & 6.1 & 39.6 & 3.6M & 6.1\\
%EfficientDet-D2 & 768$\times$768 & 42.3 & 50.8M & 11 & 42.3 & 4.7M & 11\\
%\bottomrule
%\end{tabular}
%\caption{Comparison of mAP accuracy results on MS COCO-2017 validation set for different EfficientDet models \cite{efdet_paper}.}
%\label{effdet_tab}
%\end{table}

\subsubsection{Semantic Segmentation}
We evaluate TSE on semantic segmentation trained on Cityscapes \cite{cityscapes_paper}. 
We use MobileNetV3 with an LR-ASPP segmentation head \cite{mbv3_paper} as our comparison model. We conduct the experiments with metric mIoU \cite{miou_paper}, and only exploit the 'fine' annotations in the Cityscapes dataset. Models are evaluated with a single-scale of 1024$\times$2048 input on the Cityscapes validation set. Results are shown in Table \ref{mbv3_tab}. As for object detection, we note both the very high resolution which this network operates and the spatially-sensitive nature of the task. We see that using strip pooling of only seven rows is still enough to capture the required spatial-context for channel attention.

\begin{table}
\scriptsize
\centering
\begin{tabular}{@{}llll@{}}
\toprule
 & & MobileNetV3-L & MobileNetV3-S \\
\midrule
SE & mIoU & 69.54 & 64.11 \\
   & Buffer & 36.17M & 24.51M \\
   & GFLOPs & 68.6 & 33.5 \\
\midrule
TSE$_{7 \times W}$ & mIoU & 69.2 & 63.83 \\
   & Buffer & 1.59M & 1.04M \\
   & GFLOPs & 68.6 & 33.5 \\
\midrule
TSE$_{7 \times W}^{aw}$ & mIoU & 69.02 & 63.83 \\
   & Buffer & 1.59M & 1.04M \\
   & GFLOPs & 68.6 & 33.5 \\
\bottomrule
\end{tabular}
\caption{Comparison of mIoU accuracy results on Cityscapes validation set for different MobileNetV3 models \cite{mbv3_paper}.}
\label{mbv3_tab}
\end{table}

%-------------------------------------------------------------------------
%----------------------- Conclusion --------------------------------------
%-------------------------------------------------------------------------

\section{Conclusion}
We have presented the tiled squeeze-and-excite (TSE), a new framework for channel attention that relies on non-overlapping tiles with limited spatial extent. Through analysis and comprehensive experimentation, we have shown that channel-attention learned with local spatial context is equal in performance to attention learned with global-context. We further showed that for large tiles the local-context is a good enough estimator of the global context and therefore TSE can replace SE post-training. Furthermore, TSE significantly reduces the pipeline-buffering requirements in dataflow AI accelerators while preserving baseline accuracy. We hope that our analysis and results will be an important step towards quantifying the importance of spatial context for other attention mechanisms in the future.

% Most notably, recently there has been much interest in the use of Transformers\cite{attention_is_all_you_need_paper} for vision application\cite{detr_paper,vit_paper,botnet_paper} and we hope that our work can be a step towards making the self-attention operation in vision applications more light-weight by relying on local-context only.

%-------------------------------------------------------------------------
%-------------------------------------------------------------------------
%----------------------------Bib -----------------------------------------
%-------------------------------------------------------------------------
%-------------------------------------------------------------------------

{\small
\bibliographystyle{ieee_fullname}
\bibliography{sebib}
}

%-------------------------------------------------------------------------
%-------------------------------------------------------------------------
%------------------------ Appendix1 --------------------------------------
%-------------------------------------------------------------------------
%-------------------------------------------------------------------------

% \clearpage
\clearpage
\onecolumn
% \begin{center}
% \textbf{\large Appendix: Tiled Squeeze-and-Excite}
% \end{center}
%%%%%%%%%% Merge with supplemental materials %%%%%%%%%%
%%%%%%%%%% Prefix a "S" to all equations, figures, tables and reset the counter %%%%%%%%%%

\setcounter{figure}{0}
\setcounter{table}{0}
\renewcommand{\thetable}{\Alph{section}\arabic{table}}
\renewcommand{\thefigure}{\Alph{section}\arabic{figure}}

\renewcommand{\appendixpagename}{}

\vskip .375in
   \begin{center}
      {\Large \bf Appendix: Tiled Squeeze-and-Excite \par}
   \end{center}
   
\begin{appendices}
\section{TSE Code}
An implementation in PyTorch of the TSE block is given in Figure \ref{pytorch_code_fig}.

\begin{figure}[h]
\begin{lstlisting}[language=Python]
def TSE(x, kernel, se_ratio):
    # x: input feature map [N, C, H, W]
    # kernel: tile size (Kh, Kw)
    # se_ratio: SE channel reduction ratio

    N, C, H, W = x.size()

    # tiled squeeze
    sq = nn.AvgPool2d(kernel, stride=kernel, ceil_mode=True)
    # original se excitation
    ex = nn.Sequential(
        nn.Conv2d(C, C // se_ratio, 1),
        nn.ReLU(inplace=True),
        nn.Conv2d(C // se_ratio, C, 1),
        nn.Sigmoid()
    )
    y = ex(sq(x))
    # nearest neighbor interpolation
    y = torch.repeat_interleave(y, kernel[0], dim=-2)[:,:,:H,:]
    y = torch.repeat_interleave(y, kernel[1], dim=-1)[:,:,:,:W]
    return x * y
\end{lstlisting}
\caption{PyTorch code of our TSE block}
\label{pytorch_code_fig}
\end{figure}

%-------------------------------------------------------------------------
%-------------------------------------------------------------------------
%------------------------ Appendix2 --------------------------------------
%-------------------------------------------------------------------------
%-------------------------------------------------------------------------

\section{Different Design Selections}
Here, we introduce different design selections for TSE that minimize the spatial context of the operation without losing accuracy. We will focus on two different modifications: (a) change the channel reduction ratio, and (b) change the 1x1 convolutions of the excitation step to different operations, \eg, 3x3 convolution. We note that those alteration change the number of parameters in the model and thus make it incompatible with standard SE block. The target of this experiment is to examine whether a model with single row strip pooling or without any pooling can achieve the same accuracy as SE network with GAP.

\begin{table}[b]
\setlength\tabcolsep{3pt}
\scriptsize
\centering
\begin{tabular}{@{}lllll@{}}
\toprule
Method & Params & GFLOPs & Buffer & Top-1 \\
\midrule
SE & 6.2M & 0.79 & 1.07M & 76.30 \\
\midrule
TSE$_{1\times W}C_{1\times 1}R_4$ & 6.2M & 0.80 & 0.06M & 75.79 \\
TSE$_{1\times W}C_{1\times 1}R_2$ & 7.08M & 0.81 & 0.06M & 75.90 \\
TSE$_{1\times W}C_{3\times 1}R_4$ & 7.8M & 0.82 & 0.06M & 76.40 \\
TSE$_{1\times W}C_{3\times 1}R_2$ & 10.3M & 0.84 & 0.06M & 76.43 \\
\midrule
TSE$_{1\times 1}C_{1\times 1}R_4$ & 6.2M & 0.93 & - & 75.54 \\
TSE$_{1\times 1}C_{1\times 1}R_2$ & 7.08M & 2.84 & - & 75.90 \\
%TSE$_{1,1}$-$3x3$-$4$ & 12.8M & 1.82 & - & 76.90 \\
TSE$_{1\times 1}C_{3\times 3}R_2$ & 20.17M & 2.84 & - & 77.14 \\
\bottomrule
\end{tabular}
\caption{Different design selections for TSE. Top-1 accuracy comparison of different RegNetY-800MF models on ImageNet-1K. The naming convention is: TSE$_{h\times w}C_{k_x\times k_y}R_c$ where $C_{k_x\times k_y}$ and $R_c$ are the dimensions of the convolution kernel and the channel reduction ratio, respectively.}
\label{wse_tab}
\end{table}

\par We consider two operations to swap the 1x1 convolution with: $Conv2D_{3\times 3}$ and $Conv2D_{3\times 1}$ and each model is named according to the following scheme: TSE$_{h \times w}C_{k_x\times k_y}R_c$ where $C_{k_x\times k_y}$ and $R_c$ are the dimensions of the convolution kernel and the channel reduction ratio, respectively. For instance, TSE$_{1 \times W}C_{3\times 1}R_2$ is a model with strip pooling of a single row, channel reduction ratio of $2$ and two $Conv2D_{3\times 1}$ operations. The baseline in our experiment is RegNetY-800MF model \cite{regnet_paper} which has a native channel reduction ratio of $4$. The results are shown in Table \ref{wse_tab}.

\begin{table}
\setlength\tabcolsep{2pt}
\scriptsize
\centering
\begin{tabular}{@{}llllllll@{}}
\toprule
Model & Input & \multicolumn{3}{c}{SE} & \multicolumn{3}{c}{TSE$_{1 \times W}$} \\
& & mAP & Buffer & GFLOPs & mAP & Buffer & GFLOPs \\
\midrule
EfficientDet-D0 & 512$\times$512 & 33.8 & 13.8M & 2.5 & 34.4 & 0.28M & 2.7 \\
EfficientDet-D1 & 640$\times$640 & 39.0 & 33.6M & 6.1 & 39.5 & 0.52M & 6.7 \\
EfficientDet-D2 & 768$\times$768 & 42.3 & 50.8M & 11 & 42.4 & 0.68M & 11.8 \\
\bottomrule
\end{tabular}
\caption{Comparison of mAP accuracy results on MS COCO-2017 validation set for different EfficientDet models \cite{efdet_paper} with TSE$_{1\times w}C_{3\times 1}R_4$.}
\label{efdet_wse_tab}
\end{table}

\par\textbf{TSE Without Pooling.} In the first experiment we use TSE without pooling, e.g., TSE$_{1\times 1}$ (bottom part of Table \ref{wse_tab}). The baseline top-1 accuracy of 75.54\% obtained with the original excitation step of SENet and without any pooling. First, we show that decreasing the channel reduction ratio can increase the accuracy of the network by 0.36\%. However, even with reduction ratio of $2$ the network has a large accuracy margin compare to the SE model. Second, we replace the $Conv2D_{1\times 1}$ operation with $Conv2D_{3\times 3}$. This replacement increases the number of parameters and compute but also improves network accuracy above the SE baseline. The purpose of this experiment is to show that GAP (or any global spatial context) is not mandatory for channel attention and variations in the excitation step can 'overcome' the lack of spatial context and even give improvement over more basic excitation schemes with global pooling. It also shows the trade-off between the number of parameters and compute to pipeline buffering which optimally can be optimized to a specific hardware.

\par\textbf{Single Row Strip-tiling.} Here, we use TSE with a single row strip-tiling, \eg, TSE$_{1\times W}$. Unlike the first experiment, here, we squeeze the rows which induces less computation. We also note that changing the channel reduction ratio has limited accuracy gains. To match SE accuracy in this case, we swap the $Conv2D_{1\times 1}$ convolution to $Conv2D_{3\times 1}$ which increases the number of parameters (even with the same channel reduction ratio). This variant of TSE shows how minimum amount of spatial context can be enough to generate meaningful attention factors. We further verified that TSE$_{1\times W}C_{3\times 1}R_4$  can be used with high resolution networks as well. For this experiment, we employ the EfficientDet \cite{efdet_paper} models. The results are shown in Table \ref{efdet_wse_tab} and shows that the same accuracy can be achieved with less pipeline buffering with the cost of adding a small amount of additional computations.

\par In summary, both experiments show that there is a rather large design space for SE-like operations. Spatial squeezing has a dual role of decreasing computation and aggregating spatial context, however, the more spatial information is squeezed, the greater the buffering required. At the extreme, spatial squeezing can be avoided altogether with the cost of increased computations and reduced accuracy. Adding a spatial component to the excite operation improves the performance of channel attention while significantly increasing the number of parameters. In fact, using $Conv2D_{3\times 3}$ has superior performance than the original SE block. The original SE block uses GAP to reduce the compute to minimum and a simple excitation to reduce the parameter count to a minimum, however, is also maximizes the required pipeline buffering. Based on the above observations, TSE is designed to bring all three of compute, parameters and buffering to a minimum while maintaining accuracy.

\end{appendices}

\end{document}